\documentclass[a4paper,fleqn]{cas-dc}

\usepackage[authoryear,longnamesfirst]{natbib}
\usepackage{caption}
\usepackage{placeins} 

\def\tsc#1{\csdef{#1}{\textsc{\lowercase{#1}}\xspace}}
\tsc{WGM}
\tsc{QE}
\tsc{EP}
\tsc{PMS}
\tsc{BEC}
\tsc{DE}

\makeatletter
\@ifundefined{orcid}{}{\renewcommand{\orcid}[1]{}}%
\@ifundefined{orcidlink}{}{\renewcommand{\orcidlink}[1]{}}%
\@ifundefined{printorcid}{}{}%
\@ifundefined{orcidname}{}{}%
\makeatother

\begin{document}
\let\WriteBookmarks\relax
\def\floatpagepagefraction{1}
\def\textpagefraction{.001}

\shorttitle{Rotational motion-induced error compensation algorithm}

\shortauthors{Seong-Jin An et~al.}

\title [mode = title]{Rotational Motion-Induced Error Compensation for Phase-Shifting Profilometry-Based Eye Reconstruction}                      

%
\author[1]{Seong-Jin An}
\credit{Writing -- original draft, Methodology, Software, Investigation, Formal analysis, Validation}
\affiliation[1]{organization={Department of Mechanical Engineering, Yonsei University},
    addressline={03722}, 
    city={Seoul},
    country={Republic of Korea}}

\author[1]{Sanghoon Jeon}
\credit{Conceptualization, Methodology, Software}

\author[2]{Yatong An}
\ead{yatong@meta.com}
\cormark[1]
\credit{Writing – review \& editing, Validation, Supervision, Conceptualization, Funding acquisition}
\affiliation[2]{organization={Meta Reality Labs},
    city={Redmond},
    postcode={98052},
    state={Washington},
    country={USA}}

\author[1]{Jae-Sang Hyun}
\ead{hyun.jaesang@yonsei.ac.kr}
\cormark[1]
\credit{Writing – review \& editing, Validation, Supervision, Funding acquisition, Project administration}

\cortext[cor1]{Corresponding authors}

\begin{abstract}
With the proliferation of immersive Head-Mounted Displays (HMDs) for Virtual and Augmented Reality (VR/AR), reliable and high-precision eye tracking has become increasingly important. Conventional 2D image-based methods offer low system complexity but remain limited in stability, accuracy, and robustness. Three-dimensional ocular surface reconstruction can provide richer geometric information, and structured light profilometry is particularly attractive because it enables dense and accurate surface measurement. However, Phase-Shifting Profilometry (PSP), which estimates phase from sequentially acquired fringe images, is highly susceptible to motion-induced errors when the eye rotates between frames. This study proposes a rotational motion compensation framework for PSP-based dynamic 3D eye reconstruction. Relative eye rotation is estimated from image-based motion cues using a user-specific 3D eye model in a spherical-coordinate domain. The estimated motion is then used to compensate for camera-pixel mismatch and phase-shift errors caused by inter-frame rotation. A region-wise optimization strategy is further introduced to reduce residual artifacts by independently refining the compensation strength in different ocular regions. Experiments with a rotating fake eye under non-uniform motion demonstrate that the proposed method substantially suppresses motion-induced deformation and improves reconstruction accuracy. An additional experiment with a non-spherical rigid object indicates that the compensation principle is not restricted to spherical eye geometry. These results establish a practical basis for stable PSP-based dynamic 3D eye reconstruction toward future high-precision eye tracking in immersive environments.
\end{abstract}


\begin{highlights}
\item An encoder-free motion compensation framework is proposed for dynamic 3D eye reconstruction.
\item Eye rotation-induced pixel mismatch and phase-shift errors are compensated in PSP.
\item Region-wise optimization improves reconstruction robustness under low frame-rate conditions.
\end{highlights}


\maketitle

\section{Introduction}

Recent years have witnessed a transformative shift in digital paradigms through the rapid advancement of Virtual Reality (VR) and Augmented Reality (AR) technologies. Driven largely by the commercial expansion of the gaming industry, these immersive technologies have transcended entertainment to redefine a wide array of sectors, including healthcare, education, and industrial manufacturing~\citep{anthes2016state}. At the core of these experiences is the ability of Head-Mounted Displays (HMDs) to synchronize digital content with human movement. While initial developments in VR were primarily centered on perfecting head-orientation tracking to maintain immersion, the technological frontier is increasingly shifting toward the integration of sophisticated eye-tracking solutions to enable more natural and responsive interaction~\citep{clay2019eye, kar2017review}.

In this context, eye tracking has emerged as an important interaction modality for immersive environments. Unlike traditional input methods that rely on external controllers or broad physical movements, gaze-based interaction allows for a more intuitive control scheme that mirrors natural human intent~\citep{sibert2000evaluation, morimoto2005eye}. By estimating the user's point of attention, HMDs can deliver responsive interfaces and support efficient navigation. Furthermore, this technology serves as a cornerstone for system-level performance optimizations, such as foveated rendering, which concentrates computational resources near the user's gaze direction~\citep{guenter2012foveated, patney2016towards}. Consequently, eye tracking is now recognized as an important component for creating the next generation of responsive and user-centric immersive platforms.

Despite its potential, contemporary HMD-based eye tracking still faces significant challenges when it is used as a reliable interaction modality. Recent comparative studies have reported that unmodified eye tracking can suffer from higher miss rates than traditional input methods, such as hand controllers or head tracking~\citep{fernandes2023leveling}. One important limitation of conventional 2D image-based gaze estimation is that its accuracy can degrade toward the periphery of the field of view (FOV). In commercial HMDs, the highest accuracy is often achieved near the central viewing region, whereas larger gaze angles can introduce pupil foreshortening, where the pupil appears as a distorted ellipse due to oblique camera views, leading to increased gaze estimation errors~\citep{schuetz2022eye, adhanom2023eye}.

In addition, methods that rely primarily on 2D eye-image features for final gaze estimation can be sensitive to user-specific physiological characteristics and imaging conditions. Variations in eye appearance and anatomical structure may lead to inconsistent tracking quality across users~\citep{clay2019eye, nystrom2013influence}, while ambient illumination and corneal reflections can affect the reliability of extracted image features~\citep{jin2024eye}. These limitations are closely related to the inherent loss of geometric information in 2D projections. Without an explicit 3D geometric relationship between the eye and the imaging sensor, 2D feature mapping alone cannot fully account for parallax effects and optical distortions in three-dimensional space~\citep{guestrin2006general, hansen2009eye}. Therefore, there is a growing need for approaches that incorporate three-dimensional eye geometry to improve the robustness of eye tracking.

To overcome the limitations of conventional 2D eye tracking, recent studies have begun to exploit three-dimensional ocular surface information for gaze estimation. Dense 3D surface measurements of the eye can provide richer geometric cues, such as surface shape and surface normals, beyond sparse 2D image features. For example, ~\cite{wang2025accurate} demonstrated that dense 3D reconstructions of the cornea and sclera can be used for accurate eye tracking, suggesting the potential of 3D ocular surface information for robust gaze estimation.

To obtain such 3D ocular geometry, structured light profilometry is a promising approach because it can provide dense and accurate surface measurements~\citep{geng2011structured, zhang2018high}. Structured-light-based 3D eye-tracking approaches have also been explored to acquire eye-surface geometry using projected fringe patterns~\citep{zheng2023fringe}. Among structured light techniques, Phase-Shifting Profilometry (PSP) is particularly attractive because it estimates phase information from multiple phase-shifted fringe images and can achieve high-precision 3D reconstruction~\citep{zuo2018phase}. However, because PSP requires sequential image acquisition, it is vulnerable to motion-induced errors when the eye rotates between phase-shifted frames. Even small inter-frame rotations can cause camera-pixel mismatch and phase-shift variation, resulting in distorted phase maps and degraded 3D reconstruction quality.

Motion-induced errors have been investigated in conventional digital fringe projection systems. For example, ~\cite{jeon2024motion} reduced motion-induced errors in a motorized fringe projection system by estimating translational displacement using an external encoder. Such an approach is effective when the object motion can be directly measured or controlled in a known mechanical coordinate system. However, dynamic eye reconstruction presents a different challenge. The dominant motion of the eye is rotational rather than translational, and it occurs around a user-specific eyeball center. In addition, the eye motion cannot be directly measured by an external encoder in practical HMD environments, and the projected fringe patterns interfere with direct pupil detection. Therefore, a compensation method that can estimate rotational eye motion from image-based cues and map this motion into both camera and projector domains is required for PSP-based dynamic eye reconstruction.

This study therefore focuses on rotational motion-induced error compensation for PSP-based 3D eye reconstruction. The proposed method estimates the relative rotational motion of the eye using a calibrated three-dimensional eye model and image-based motion cues, and represents the motion in a spherical-coordinate domain. Based on this motion estimate, both camera-pixel errors and phase-shift errors between sequential fringe images are compensated before phase calculation. Instead of evaluating final gaze-tracking accuracy, this study specifically addresses the stability and accuracy of dynamic 3D eye reconstruction, which is an important prerequisite for reliable 3D eye tracking.

The main contributions of this study are summarized as follows. First, an image-based eye rotation estimation framework is introduced for PSP-based dynamic eye reconstruction without relying on an external motion encoder. Second, the motion-induced error compensation framework is extended from encoder-based translational motion to image-cue-based rotational eye motion by mapping the estimated inter-frame rotation to camera-pixel mismatch and phase-shift error compensation. Third, a region-wise optimization strategy is introduced to reduce residual artifacts by separately refining the compensation strength for the iris and sclera regions. Finally, the proposed framework is experimentally validated using a rotating fake eye under various frame-rate conditions, and an additional non-spherical rigid-object experiment is performed to examine whether the compensation principle can be extended beyond eye-like spherical geometry.

\section{Principle}

\subsection{Conventional Three-Step Phase-Shifting Algorithm}

Phase-shifting algorithm is a widely used technique for retrieving three-dimensional surface geometry from multiple fringe images with known phase offsets~\citep{zuo2018phase}. Among various phase-shifting methods, the three-step phase-shifting algorithm is particularly attractive because it uses the minimum number of patterns required for 3D reconstruction, thereby enabling phase estimation while minimizing the acquisition time. The intensity distributions of the three fringe images in the conventional three-step phase-shifting algorithm are expressed as follows:

\begin{equation}
I_1(u_c,v_c) = I'(u_c,v_c) + I''(u_c,v_c)\cos[\phi(u_c,v_c)-\delta]
\end{equation}

\begin{equation}
I_2(u_c,v_c) = I'(u_c,v_c) + I''(u_c,v_c)\cos[\phi(u_c,v_c)]
\end{equation}

\begin{equation}
I_3(u_c,v_c) = I'(u_c,v_c) + I''(u_c,v_c)\cos[\phi(u_c,v_c)+\delta]
\end{equation}

where $(u_c,v_c)$ denotes the camera pixel coordinate, $I'(u_c,v_c)$ corresponds to the mean intensity component, $I''(u_c,v_c)$ represents the fringe modulation amplitude, $\delta$ is the phase offset, and $\phi(u_c,v_c)$ is the wrapped phase associated with the projected fringe. In the conventional three-step phase-shifting algorithm, a stationary scene is assumed during image acquisition, so that an identical camera pixel coordinate is considered for all three fringe images.

When the phase interval is set to $\delta = 2\pi/3$, the wrapped phase can be obtained from Eqs. (1)--(3) as

\begin{equation}
\phi(u_c,v_c)=
\tan^{-1}
\left[
\frac{\sqrt{3}(I_1-I_3)}
{2I_2-I_1-I_3}
\right].
\end{equation}

The wrapped phase is restricted to the range $(-\pi,\pi]$ because Eq. (4) is obtained from the four-quadrant inverse tangent function. Consequently, the calculated phase map contains $2\pi$ discontinuities. For continuous determination of the projector-pixel position, an unwrapped phase map $\Phi(u_c,v_c)$ is required, which can be obtained by applying either temporal or spatial phase unwrapping. This unwrapping step removes the phase discontinuities by assigning and adding the proper integer multiples of $2\pi$. Hence, the unwrapped phase can be expressed as

\begin{equation}
\Phi(u_c,v_c) = \phi(u_c,v_c) + 2\pi \times k(u_c,v_c)
\end{equation}

where $k(u_c,v_c)$ denotes an integer fringe order.

\subsection{Eye Rotation Estimation}
The motion parameters used for camera-pixel and phase-shift error compensation are estimated from the rotational motion of the eye. In conventional digital fringe projection systems, motion-induced errors have been compensated by measuring translational displacement using an external encoder attached to a motorized stage~\citep{jeon2024motion}. However, the rotational motion of the eye cannot be directly measured in the same manner. Therefore, a user-specific three-dimensional eye model is constructed, and the gaze-vector variation between sequential fringe images is used to estimate the relative motion in the spherical-coordinate domain.

First, a user calibration procedure is performed under static conditions. The eye is scanned or imaged at multiple gaze poses, and the corresponding gaze vectors are extracted. Since the eye can be approximated as rotating around a fixed center, the collected gaze vectors are used to estimate the user-specific eyeball center. This center is then used as the origin of the spherical-coordinate representation. Accordingly, a surface point on the eye can be represented by its radial distance $r$ from the eyeball center and two angular coordinates, $\theta$ and $\varphi$.

For dynamic measurements, direct pupil detection from the original fringe images is unreliable because the projected sinusoidal patterns overlap with the pupil region. To suppress the fringe pattern, adjacent phase-shifted fringe images are averaged to generate a pattern-reduced image for each acquired frame. Under the assumption that the image acquisition speed is sufficiently high, the eye motion within the averaging window is small, while the phase-shifted fringe components are largely canceled. The extracted two-dimensional pupil center is then projected onto the user-specific 3D eye model, and the gaze vector is obtained by connecting the estimated eyeball center and the projected pupil center. The relative rotational motion of the eye is estimated from the temporal variation of these frame-wise gaze vectors. In this study, the estimated gaze vector is used as an intermediate motion cue for compensating inter-frame rotational displacement, rather than as a final eye-tracking output.

For the three-step phase calculation, the relative motion parameters are defined with respect to the reference fringe image used for phase computation. The neighboring fringe images are compensated and aligned to the coordinate system of the reference image. Accordingly, the motion parameters between the first and reference images and between the third and reference images are defined as
\begin{equation}
    \Delta \mathbf{q}_{12} =
    \begin{bmatrix}
    \Delta r_{12} & \Delta \theta_{12} & \Delta \varphi_{12}
    \end{bmatrix}^{T},
\end{equation}
\begin{equation}
    \Delta \mathbf{q}_{32} =
    \begin{bmatrix}
    \Delta r_{32} & \Delta \theta_{32} & \Delta \varphi_{32}
    \end{bmatrix}^{T},
\end{equation}
where $\Delta r$, $\Delta \theta$, and $\Delta \varphi$ denote the relative changes in the radial and angular components of the spherical-coordinate representation.

Since the eyeball is modeled as a rigid sphere with a fixed radius, the radial variation is assumed to be negligible. Therefore, the radial components are set to
\begin{equation}
    \Delta r_{12} = \Delta r_{32} = 0.
\end{equation}
Under this assumption, the motion parameters are determined only by the angular variations, $\Delta \theta$ and $\Delta \varphi$, which are estimated from the variation of the frame-wise gaze vectors.

These estimated motion parameters are subsequently used to compute the displacement of the corresponding camera pixel and projector phase between sequential fringe images. Specifically, $\Delta \mathbf{q}_{12}$ is used to compensate the first fringe image with respect to the reference image, whereas $\Delta \mathbf{q}_{32}$ is used to compensate the third fringe image with respect to the reference image.

\begin{figure}
    \centering
    \includegraphics[width=1\linewidth]{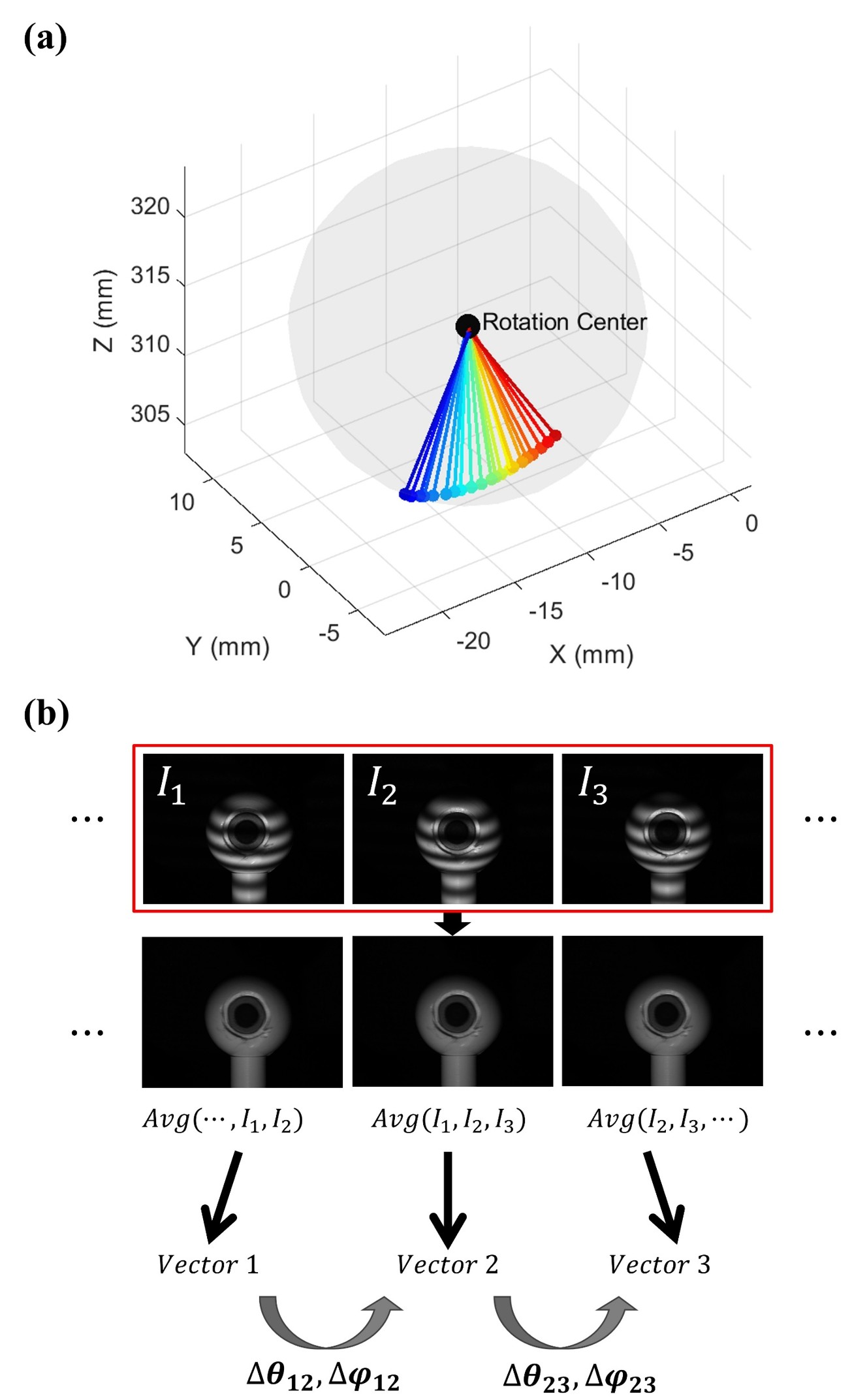}
    \caption{Eye rotation estimation process.
    (a) Estimated rotation center and spherical eye model obtained from multiple static gaze poses.
    (b) Estimation of relative eye rotation between sequential phase-shifted images using the calibrated 3D eye model.}
    \label{fig:eye_rotation}
\end{figure}

\subsection{Camera Pixel Error Compensation}
Motion during fringe pattern acquisition causes the locations of corresponding camera pixels to change across sequential fringe images. Because Eqs. (1)--(3) assume a common camera pixel location for all fringe patterns, this mismatch introduces estimation errors and degrades 3D reconstruction quality. This error is referred to as the camera pixel error. The compensation concept follows the motion-induced error formulation used in digital fringe projection systems~\citep{jeon2024motion}, but is reformulated here for rotational eye motion represented in the spherical-coordinate domain. To compensate for this error in the three-step phase-shifting algorithm, the corresponding pixel locations of the first and third fringe images are adjusted to match the pixel location of the second fringe image along the horizontal and vertical directions of the image plane. The second fringe image is chosen as the reference frame because it is captured at the temporal center of the sequence, making it the most representative of the object state and the most overlapped with the neighboring frames. Accordingly, the compensated camera pixel coordinates, instead of the common pixel location $(u_c,v_c)$, are defined as follows:
\begin{align}
u^c_{1} = u^c - \epsilon^{uc}_{12}(u^c, v^c), \qquad
v^c_{1} = v^c - \epsilon^{vc}_{12}(u^c, v^c)\\
u^c_{2} = u^c, \qquad v^c_{2} = v^c\\
u^c_{3} = u^c + \epsilon^{uc}_{32}(u^c, v^c), \qquad
v^c_{3} = v^c + \epsilon^{vc}_{32}(u^c, v^c)
\end{align}
In this formulation, $u^c_1$ and $v^c_1$ denote the corrected horizontal and vertical camera-pixel coordinates of the first fringe image, respectively, while $\epsilon^{uc}_{12}$ and $\epsilon^{vc}_{12}$ represent the relative camera-pixel errors between the first and second fringe images. The sign of the camera-pixel error can vary depending on how the principal axis of the camera is defined. When the compensated pixel coordinates lie outside the image domain, those pixels are regarded as invalid and are not used in the subsequent process.

The camera-pixel error terms are obtained from the camera projection model. In this study, the three-dimensional position of a surface point is expressed using a spherical-coordinate representation centered at the eyeball center. Based on this formulation, the image coordinates of the point can be written as
\begin{equation}
\begin{aligned}
s^{c}
\begin{bmatrix}
u^{c} \\
v^{c} \\
1
\end{bmatrix}
&=
A^{c}
\begin{bmatrix}
R^{c} & t^{c}
\end{bmatrix}
\begin{bmatrix}
x^{w} \\
y^{w} \\
z^{w} \\
1
\end{bmatrix} \\
&=
\begin{bmatrix}
M_{11} & M_{12} & M_{13} & M_{14} \\
M_{21} & M_{22} & M_{23} & M_{24} \\
M_{31} & M_{32} & M_{33} & M_{34}
\end{bmatrix}
\begin{bmatrix}
C^{w}_{x} + r \sin\theta \cos\varphi \\
C^{w}_{y} + r \sin\theta \sin\varphi \\
C^{w}_{z} + r \cos\theta \\
1
\end{bmatrix}
\end{aligned}
\label{eq:camproj_spherical}
\end{equation}
where $s^c$ is the projective scaling factor, $A^c$ is the intrinsic camera matrix, and $R^c$ and $t^c$ denote the rotation matrix and translation vector from the world coordinate system to the camera coordinate system, respectively. For notational convenience, the combined projection matrix $A^c [R^c \; t^c]$ is represented by $M$. Here, $(C_x^w, C_y^w, C_z^w)$ denotes the eyeball center in the world coordinate system, $r$ is the radial distance from the eyeball center to the target surface point, and $\theta$ and $\varphi$ are the spherical-coordinate angles.

To determine how the image-plane position changes with object motion, the projection model is differentiated with respect to the spherical-coordinate variables. For example, the sensitivity of the horizontal camera-pixel coordinate to a radial change is expressed as
\begin{equation}
\begin{split}
\frac{\partial u^c}{\partial r}
&=
\frac{
\left(M_{11}\sin\theta\cos\varphi + M_{12}\sin\theta\sin\varphi + M_{13}\cos\theta\right)s^c
}{
(s^c)^2
}
\\
&\quad
-
\frac{
\left(M_{31}\sin\theta\cos\varphi + M_{32}\sin\theta\sin\varphi + M_{33}\cos\theta\right)s^c u^c
}{
(s^c)^2
}
\end{split}
\label{eq:ducdr}
\end{equation}
This term describes the local change in the horizontal pixel coordinate caused by a small radial displacement of the surface point. In the same manner, the corresponding derivatives for the vertical direction and the angular variables can also be derived and used to estimate the pixel displacement on the image plane.

According to the spherical-coordinate formulation in Eq.~(\ref{eq:camproj_spherical}), the camera-pixel error can be approximated from the first-order change of the projected pixel coordinate with respect to the spherical variables. Unlike the previous method, which estimates the displacement in the world coordinate system using the motor encoder of a linear stage, the proposed method computes the camera-pixel error from the motion parameters defined in the spherical-coordinate domain. Accordingly, the horizontal camera-pixel error terms between adjacent fringe images are expressed as
\begin{align}
\epsilon^{uc}_{12}(u^c,v^c) &= \mathrm{round}\!\left[
\frac{\partial u^c}{\partial r}\Delta r_{12}
+
\frac{\partial u^c}{\partial \theta}\Delta\theta_{12}
+
\frac{\partial u^c}{\partial \varphi}\Delta\varphi_{12}
\right]\\
\epsilon^{uc}_{32}(u^c,v^c) &= \mathrm{round}\!\left[
\frac{\partial u^c}{\partial r}\Delta r_{32}
+
\frac{\partial u^c}{\partial \theta}\Delta\theta_{32}
+
\frac{\partial u^c}{\partial \varphi}\Delta\varphi_{32}
\right]
\end{align}
Here, $\mathrm{round}[\cdot]$ denotes the rounding operator that converts the estimated sub-pixel displacement into the nearest integer pixel value. The motion parameters $\Delta r$, $\Delta \theta$, and $\Delta \varphi$ are obtained from the eye rotation estimation described in Section 2.2.

Using the camera-pixel error terms defined above, the motion-induced mismatch among the three fringe images can be compensated. Under this compensation, the background intensity, fringe modulation, and wrapped phase corresponding to the same surface location are approximately matched across the three frames. Thus, the relationships in (\ref{eq:camproj_spherical}) can be written as
{\setlength{\abovedisplayskip}{4pt}
\setlength{\belowdisplayskip}{4pt}
\setlength{\abovedisplayshortskip}{4pt}
\setlength{\belowdisplayshortskip}{4pt}
\begin{equation}
\left\{
\begin{aligned}
I'_1(u^c_1,v^c_1) &\approx I'_2(u^c,v^c)=I'(u^c,v^c)\approx I'_3(u^c_3,v^c_3)\\
I''_1(u^c_1,v^c_1) &\approx I''_2(u^c,v^c)=I''(u^c,v^c)\approx I''_3(u^c_3,v^c_3)\\
\phi_1(u^c_1,v^c_1) &\approx \phi_2(u^c,v^c)=\phi(u^c,v^c)\approx \phi_3(u^c_3,v^c_3)
\end{aligned}
\right.
\label{eq:matched_terms}
\end{equation}
}

Furthermore, when the camera-pixel error and the phase-shift variation are both taken into account, the general expressions of the three fringe images under dynamic motion can be described as
{\setlength{\abovedisplayskip}{4pt}
\setlength{\belowdisplayskip}{4pt}
\setlength{\abovedisplayshortskip}{4pt}
\setlength{\belowdisplayshortskip}{4pt}
\begin{align}
\begin{split}
I_1(u^c_1,v^c_1)
&=
I'_1(u^c_1,v^c_1) \\
&\quad + I''_1(u^c_1,v^c_1)
\cos\!\left[\phi_1(u^c_1,v^c_1)-\delta_1(u^c,v^c)\right],
\end{split}\\
I_2(u^c,v^c)
&=
I'_2(u^c,v^c)
+
I''_2(u^c,v^c)
\cos\!\left[\phi_2(u^c,v^c)\right],\\
\begin{split}
I_3(u^c_3,v^c_3)
&=
I'_3(u^c_3,v^c_3) \\
&\quad + I''_3(u^c_3,v^c_3)
\cos\!\left[\phi_3(u^c_3,v^c_3)+\delta_3(u^c,v^c)\right].
\end{split}
\end{align}
}
Here, $\delta_1(u^c,v^c)$ and $\delta_3(u^c,v^c)$ denote the corrected phase shifts of the first and third fringe images, respectively, allowing for the fact that the motion-induced phase variation is not spatially constant.

By substituting the relationships in (\ref{eq:matched_terms}) into the above equations, the fringe image model can be simplified as follows:
{\setlength{\abovedisplayskip}{4pt}
\setlength{\belowdisplayskip}{4pt}
\setlength{\abovedisplayshortskip}{4pt}
\setlength{\belowdisplayshortskip}{4pt}
\begin{align}
\begin{split}
I_1(u^c_1,v^c_1)
&=
I'(u^c,v^c) \\
&\quad + I''(u^c,v^c)
\cos\!\left[\phi(u^c,v^c)-\delta_1(u^c,v^c)\right],
\end{split}\\
I_2(u^c,v^c)
&=
I'(u^c,v^c)
+
I''(u^c,v^c)
\cos\!\left[\phi(u^c,v^c)\right],\\
\begin{split}
I_3(u^c_3,v^c_3)
&=
I'(u^c,v^c) \\
&\quad + I''(u^c,v^c)
\cos\!\left[\phi(u^c,v^c)+\delta_3(u^c,v^c)\right].
\end{split}
\end{align}
}
Compared with the original formulation, the number of unknowns is substantially reduced in the simplified model.

\subsection{Phase Shift Error Compensation}
In addition to camera-pixel mismatch, motion during phase-shifted image acquisition also changes the effective phase interval of the projected fringe pattern. This phase-shift error has been considered in previous motion-induced error compensation for digital fringe projection~\citep{jeon2024motion}; in this study, the same compensation principle is extended to rotational eye motion by estimating the phase variation in the projector domain from spherical-coordinate motion parameters. Since Eq. (4) is derived under the assumption of a constant phase shift, the phase terms should be reformulated by incorporating phase-shift compensation terms. Accordingly, the corrected phase shifts are expressed as
{\setlength{\abovedisplayskip}{4pt}
\setlength{\belowdisplayskip}{4pt}
\setlength{\abovedisplayshortskip}{4pt}
\setlength{\belowdisplayshortskip}{4pt}
\begin{align}
\delta_1(u^c,v^c) &= \frac{2\pi}{3} - \epsilon^{up}_{12}(u^c,v^c),\\
\delta_3(u^c,v^c) &= \frac{2\pi}{3} + \epsilon^{up}_{32}(u^c,v^c).
\end{align}
}
Here, $\epsilon^{up}_{12}(u^c,v^c)$ and $\epsilon^{up}_{32}(u^c,v^c)$ denote the phase-shift errors between the first and second fringe images and between the third and second fringe images, respectively.

The phase-shift error is estimated in a manner similar to the camera-pixel error, except that the projector pinhole model and phase units are used instead of the camera model and pixel units. If the fringe pattern varies only along the $u^p$ direction of the projector, the phase-shift errors can be written as
{\setlength{\abovedisplayskip}{4pt}
\setlength{\belowdisplayskip}{4pt}
\setlength{\abovedisplayshortskip}{4pt}
\setlength{\belowdisplayshortskip}{4pt}
\begin{align}
\epsilon^{up}_{12}(u^c,v^c)
&=
\frac{2\pi}{\lambda}
\left(
\frac{\partial u^p}{\partial r}\Delta r_{12}
+
\frac{\partial u^p}{\partial \theta}\Delta \theta_{12}
+
\frac{\partial u^p}{\partial \varphi}\Delta \varphi_{12}
\right),\\
\epsilon^{up}_{32}(u^c,v^c)
&=
\frac{2\pi}{\lambda}
\left(
\frac{\partial u^p}{\partial r}\Delta r_{32}
+
\frac{\partial u^p}{\partial \theta}\Delta \theta_{32}
+
\frac{\partial u^p}{\partial \varphi}\Delta \varphi_{32}
\right).
\end{align}
}
where $\lambda$ denotes the fringe pitch used to convert projector-pixel displacement into phase units.

Furthermore, by substituting Eqs.~(23) and (24) into Eqs.~(20)--(22) and solving them simultaneously, the corrected wrapped phase can be obtained as
{\setlength{\abovedisplayskip}{4pt}
\setlength{\belowdisplayskip}{4pt}
\setlength{\abovedisplayshortskip}{4pt}
\setlength{\belowdisplayshortskip}{4pt}
\begin{equation}
\phi(u^c,v^c)
=
\tan^{-1}
\left[
\frac{A_1\cos(\delta_1)+A_2\cos(\delta_3)+A_3}
{-A_1\sin(\delta_1)+A_2\sin(\delta_3)}
\right]
\end{equation}
}
with three intermediate variables as
{\setlength{\abovedisplayskip}{4pt}
\setlength{\belowdisplayskip}{4pt}
\setlength{\abovedisplayshortskip}{4pt}
\setlength{\belowdisplayshortskip}{4pt}
\begin{align}
A_1 &= I_3(u^c_3,v^c_3)-I_2(u^c,v^c),\\
A_2 &= I_2(u^c,v^c)-I_1(u^c_1,v^c_1),\\
A_3 &= I_1(u^c_1,v^c_1)-I_3(u^c_3,v^c_3).
\end{align}
}
If the motion is uniform, the phase-shift errors $\epsilon^{up}_{12}(u^c,v^c)$ and $\epsilon^{up}_{32}(u^c,v^c)$ can be assumed to be identical. In this case, the corrected wrapped phase can be simplified as
{\setlength{\abovedisplayskip}{4pt}
\setlength{\belowdisplayskip}{4pt}
\setlength{\abovedisplayshortskip}{4pt}
\setlength{\belowdisplayshortskip}{4pt}
\begin{equation}
\phi
=
\tan^{-1}
\left[
\frac{(2+\cos(\epsilon)+\sqrt{3}\sin(\epsilon))(I_1-I_3)}
{(\sqrt{3}\cos(\epsilon)-\sin(\epsilon))(2I_2-I_1-I_3)}
\right]
\end{equation}
}
where $\epsilon$ represents either $\epsilon^{up}_{12}(u^c,v^c)$ or $\epsilon^{up}_{32}(u^c,v^c)$.

\subsection{Phase Unwrapping Using Geometric Constraints}
In dynamic measurement, increasing the number of projected patterns generally aggravates motion-induced errors and reduces the effective acquisition speed. For this reason, it is desirable to minimize the number of required fringe patterns while still obtaining an unwrapped phase map. To address this issue, a pixel-wise absolute phase unwrapping strategy based on the geometric constraints of the structured light system is employed~\citep{an2016pixel}. This approach utilizes the calibrated geometric relationship between the camera and the projector to recover the absolute phase without requiring additional image acquisition.

The method is established on the pinhole models of both the camera and the projector. First, a virtual reference plane is defined at the nearest depth of interest in the world coordinate system, denoted by $z_{\min}$. For each camera pixel, the corresponding 3D coordinates on this plane can be determined from the camera projection model because the depth of the plane is known. Then, by projecting these 3D points onto the projector coordinate system, the absolute phase corresponding to each camera pixel can be estimated. This minimum absolute phase map, denoted by $\Phi_{\min}$, serves as a reference for unwrapping the wrapped phase map.

Because the virtual reference plane is placed sufficiently close to the object surface, the phase difference between the desired absolute phase $\Phi$ and the reference phase $\Phi_{\min}$ remains smaller than $2\pi$. Under this condition, the wrapped phase $\phi$ and the reference phase $\Phi_{\min}$ satisfy the following relationship:

\begin{equation}
2\pi \times (k(u_c,v_c)-1)
<
\Phi_{\min}-\phi
<
2\pi \times k(u_c,v_c),
\end{equation}

from which the fringe order can be determined as

\begin{equation}
k(u_c,v_c)
=
\left\lceil
\frac{\Phi_{\min}-\phi}{2\pi}
\right\rceil,
\end{equation}

where $\lceil \cdot \rceil$ denotes the ceiling operator.

Once the fringe order is obtained, the absolute phase can be recovered pixel by pixel from the wrapped phase map. This procedure enables phase unwrapping with only a minimal number of projected patterns, which is particularly advantageous in dynamic scenes where additional exposures would increase motion artifacts.

However, when a single fixed reference plane is used for the entire scene, the distance between the object surface and the reference plane may vary across the field of view, which can lead to local unwrapping errors. To alleviate this issue, the reference plane can be adjusted according to the geometry of the measurement setup so that the phase difference between the object and the reference remains within a valid range over the target region.

\begin{figure}
    \centering
    \includegraphics[width=0.7\linewidth]{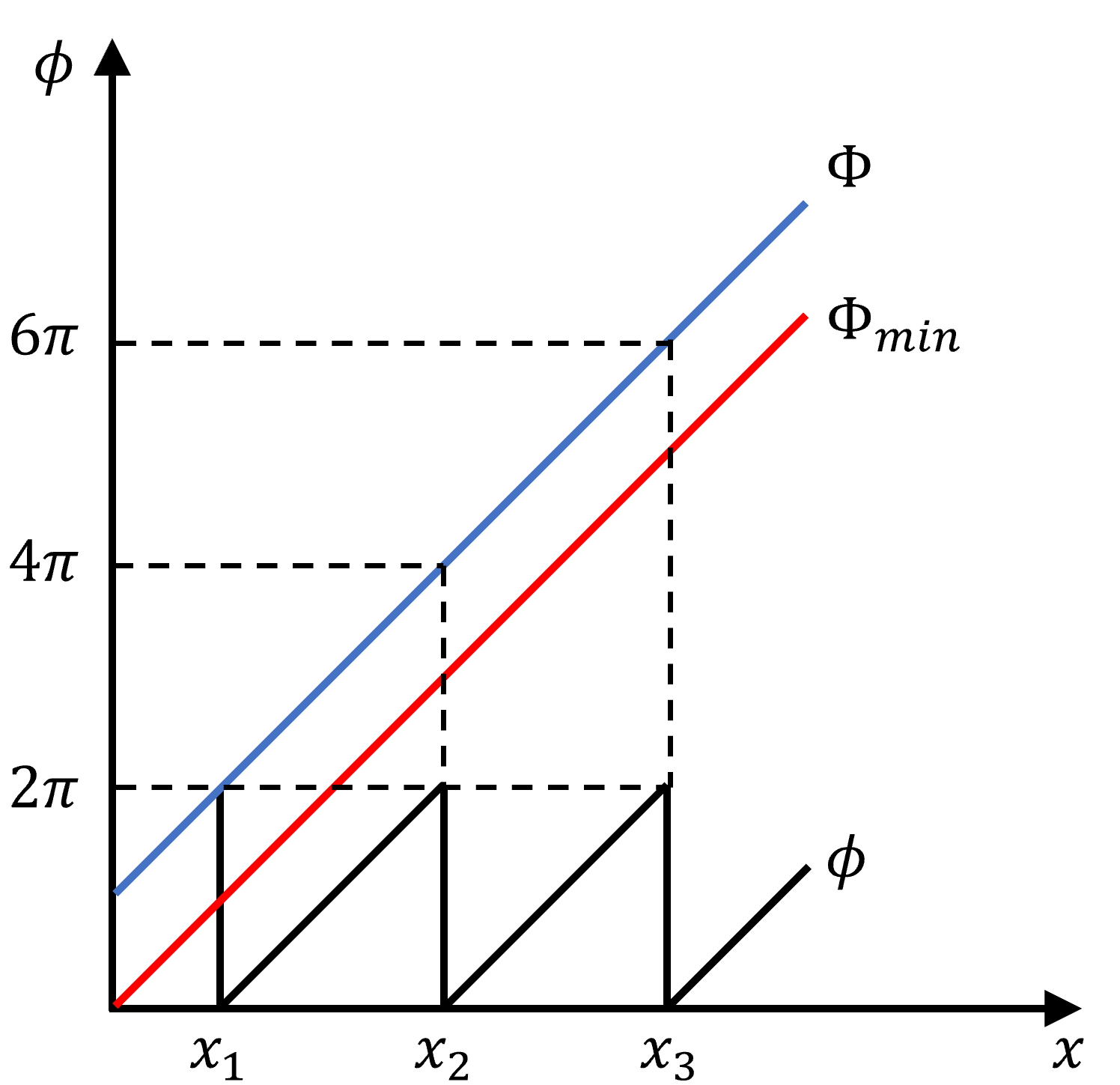}
    \caption{Conceptual principle of geometric-constraint-based phase unwrapping, where the minimum phase map is used to determine the fringe order of the wrapped phase.}
    \label{fig:geometric_constraint}
\end{figure}

\subsection{Region-Wise Optimization}
Although the proposed motion compensation model estimates camera-pixel and phase-shift errors in the spherical-coordinate domain, applying a single compensation strength to the entire ocular surface may not be optimal. Because the iris and sclera regions have different geometric and optical characteristics, residual artifacts can remain locally after motion compensation. Therefore, a region-wise optimization strategy is introduced to refine the compensation strength separately for different ocular regions.

In this study, the reconstructed eye surface is divided into the iris and sclera regions, and the estimated compensation terms are scaled independently for each region. The iris region is determined by constructing a circular iris boundary on the estimated iris plane using the eye center, pupil center, surface normal, and predefined iris radius, and then projecting this boundary onto the reference image plane. The remaining valid ocular surface outside the projected iris boundary is defined as the sclera region. To avoid abrupt changes near the region boundary, a narrow transition band is excluded or assigned an intermediate compensation scale between the iris and sclera regions.

Let $\alpha_{k,R}$ denote the compensation scale factor for inter-frame motion $k$ and region $R$, where $k \in \{12,32\}$ and $R \in \{\mathrm{iris}, \mathrm{sclera}\}$. The scaled motion parameters can be expressed as

\begin{equation}
\Delta \mathbf{q}^{*}_{k,R} = \alpha_{k,R}\Delta \mathbf{q}_{k},
\quad R \in \{\mathrm{iris}, \mathrm{sclera}\}, \quad k \in \{12,32\}.
\end{equation}

In the implementation, the scale factor is parameterized as

\begin{equation}
\alpha_{k,R}=1+\delta_{k,R},
\end{equation}

where $\delta_{k,R}$ represents the region-wise adjustment from the original compensation amount. Accordingly, four scale parameters are considered:

\begin{equation}
\boldsymbol{\alpha}
=
\left[
\alpha_{12,\mathrm{iris}},
\alpha_{32,\mathrm{iris}},
\alpha_{12,\mathrm{sclera}},
\alpha_{32,\mathrm{sclera}}
\right]^{T}.
\end{equation}

The optimal scale parameters are determined by minimizing the reconstruction error between the region-wise compensated result and a pose-aligned high-frame-rate reference reconstruction:

\begin{equation}
\boldsymbol{\alpha}^{*}
=
\arg\min_{\boldsymbol{\alpha}}
\mathrm{RMSE}
\left(
\mathbf{P}^{\mathrm{comp}}(\boldsymbol{\alpha}),
\mathbf{P}^{\mathrm{ref}}
\right),
\end{equation}

where $\mathbf{P}^{\mathrm{comp}}(\boldsymbol{\alpha})$ denotes the compensated 3D points reconstructed using the region-wise scale parameters, and $\mathbf{P}^{\mathrm{ref}}$ denotes the corresponding reference points. In this study, the pose-aligned 500 FPS compensated reconstruction is used as the reference for determining the region-wise scale parameters.

This region-wise optimization reduces residual compensation mismatch between ocular regions and improves the stability of the reconstructed 3D eye surface, especially under low-frame-rate conditions where inter-frame rotational displacement becomes large.

\section{Experiments}
\subsection{Experiment Setup}
The experimental setup was configured to validate the proposed motion compensation method under high-speed rotational eye motion. The structured-light system consisted of a CMOS camera (ORX-10G-32S4M-C) with a 16 mm camera lens (Computar M1614-MP2), and a DLP projector (Texas Instruments LightCrafter 4500). The projector and camera resolutions were $912 \times 1140$ pixels and $640 \times 480$ pixels, respectively. The image acquisition and pattern projection sequences were synchronized to operate at up to 500 Hz. This synchronization enabled high-speed acquisition of phase-shifted fringe images and reduced motion-induced inconsistencies between consecutive frames.

A fake eye was mounted on a servo motor (DYNAMIXEL XC330-M181-T) to reproduce controlled rotational motion. The no-load speed of the servo motor was specified as 129 rpm, corresponding to 774$^\circ$/s. Although the servo motor has a built-in encoder, the encoder was used only to monitor the eye-motion profile for experimental verification and was not used as an input to the proposed motion compensation algorithm. Instead, the compensation process relied on image-based motion cues, consistent with practical eye-tracking scenarios where external encoders are not available. The fake eye had a diameter of 24 mm, similar to the average diameter of the human eyeball. To isolate motion-induced errors in phase-shifting profilometry and exclude the effect of corneal refraction, the corneal part of the fake eye was removed in this experiment.

\begin{figure}
    \centering
    \includegraphics[width=1\linewidth]{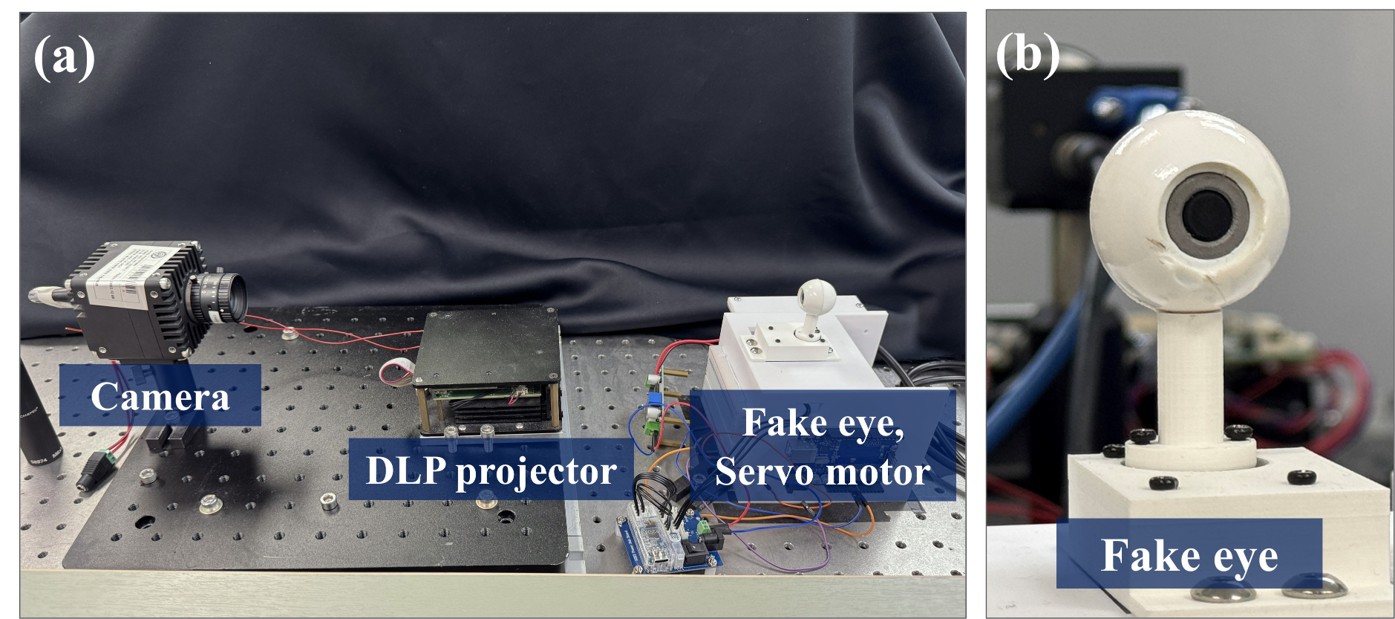}
    \caption{Experimental setup: (a) structured-light system with a fake eye and servo motor; (b) detailed view of the fake eye.}
    \label{fig:experiment_setup}
\end{figure}

\subsection{Motion-Induced Error Compensation}
The proposed motion compensation method was first evaluated under high-speed rotational eye motion. The fake eye mounted on the servo motor was driven to perform reciprocating rotation over a 60-degree range, with a maximum angular velocity of approximately $700^\circ$/s. This motion condition was designed to reproduce a challenging dynamic situation comparable to the peak rotational velocity of the human eye. The reconstruction results before and after applying the proposed motion compensation were then compared to verify the effectiveness of the algorithm.

\begin{figure}
    \centering
    \includegraphics[width=0.8\linewidth]{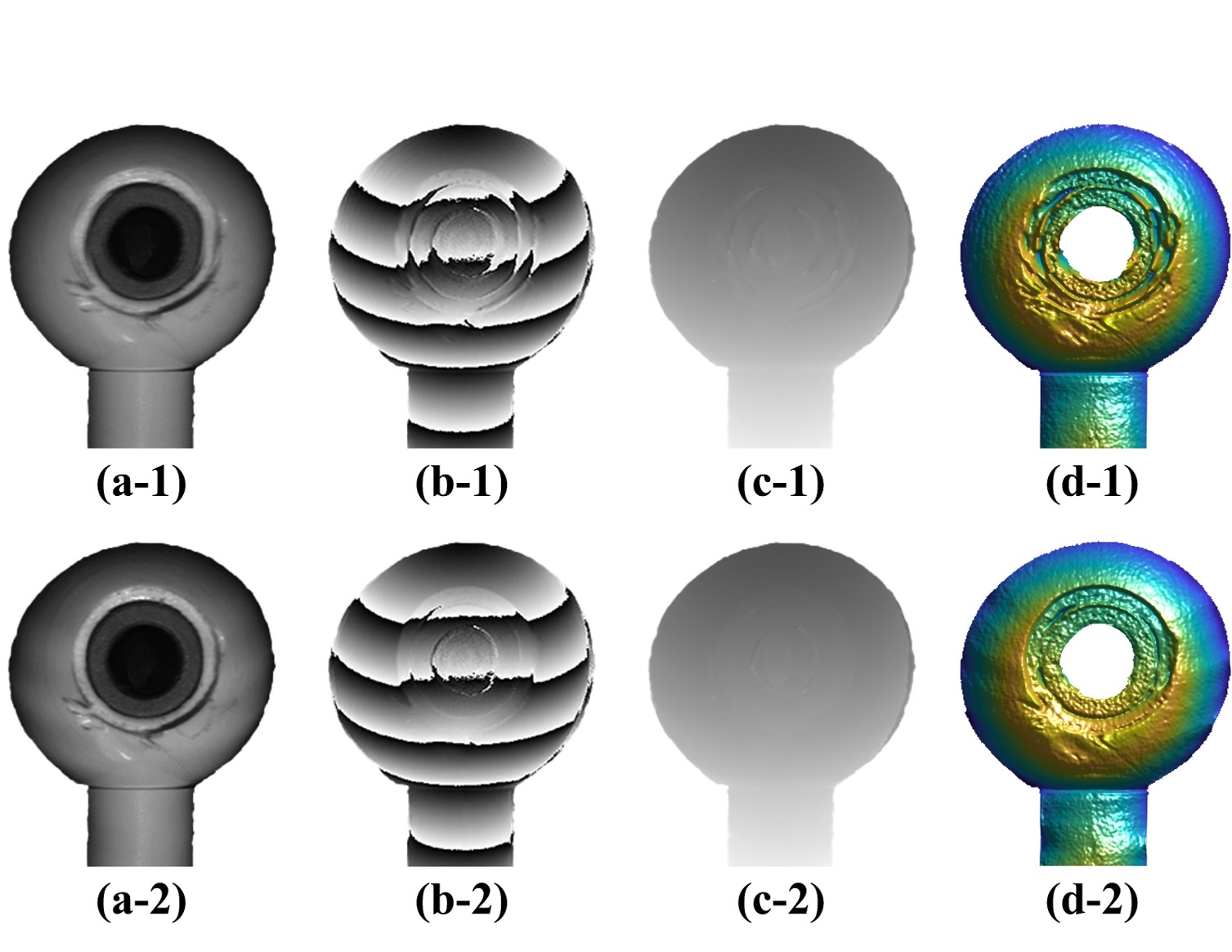}
    \caption{Effect of the proposed motion compensation on phase maps and 3D reconstruction. The first row shows the results before compensation, and the second row shows the results after compensation: (a) averaged image, (b) wrapped phase, (c) unwrapped phase, and (d) reconstructed 3D result.}
    \label{fig:Before_After_Compensation}
\end{figure}

Fig.~\ref{fig:Before_After_Compensation} shows the effect of the proposed motion compensation on the fringe images, wrapped phase, and reconstructed 3D surface. Before compensation, the three phase-shifted fringe images were not spatially aligned because the fake eye rotated during sequential image acquisition. This inter-frame mismatch distorted the estimated phase, especially in the iris region, and consequently produced severe deformation in the reconstructed 3D surface. Since accurate 3D gaze-vector estimation relies on the stable reconstruction of the iris and pupil region, this motion-induced distortion can degrade the quality of the pupil–iris surface reconstruction, which is a prerequisite for reliable 3D eye tracking.

After applying the proposed compensation method, the neighboring fringe images were aligned with respect to the reference frame by correcting both camera-pixel and phase-shift errors. As a result, the phase distortion in the iris region was significantly reduced, and the reconstructed 3D surface became more geometrically consistent. These results demonstrate that the proposed method effectively suppresses the motion-induced errors caused by eye rotation during three-step phase-shifting profilometry.

\begin{figure}
    \centering
    \includegraphics[width=1\linewidth]{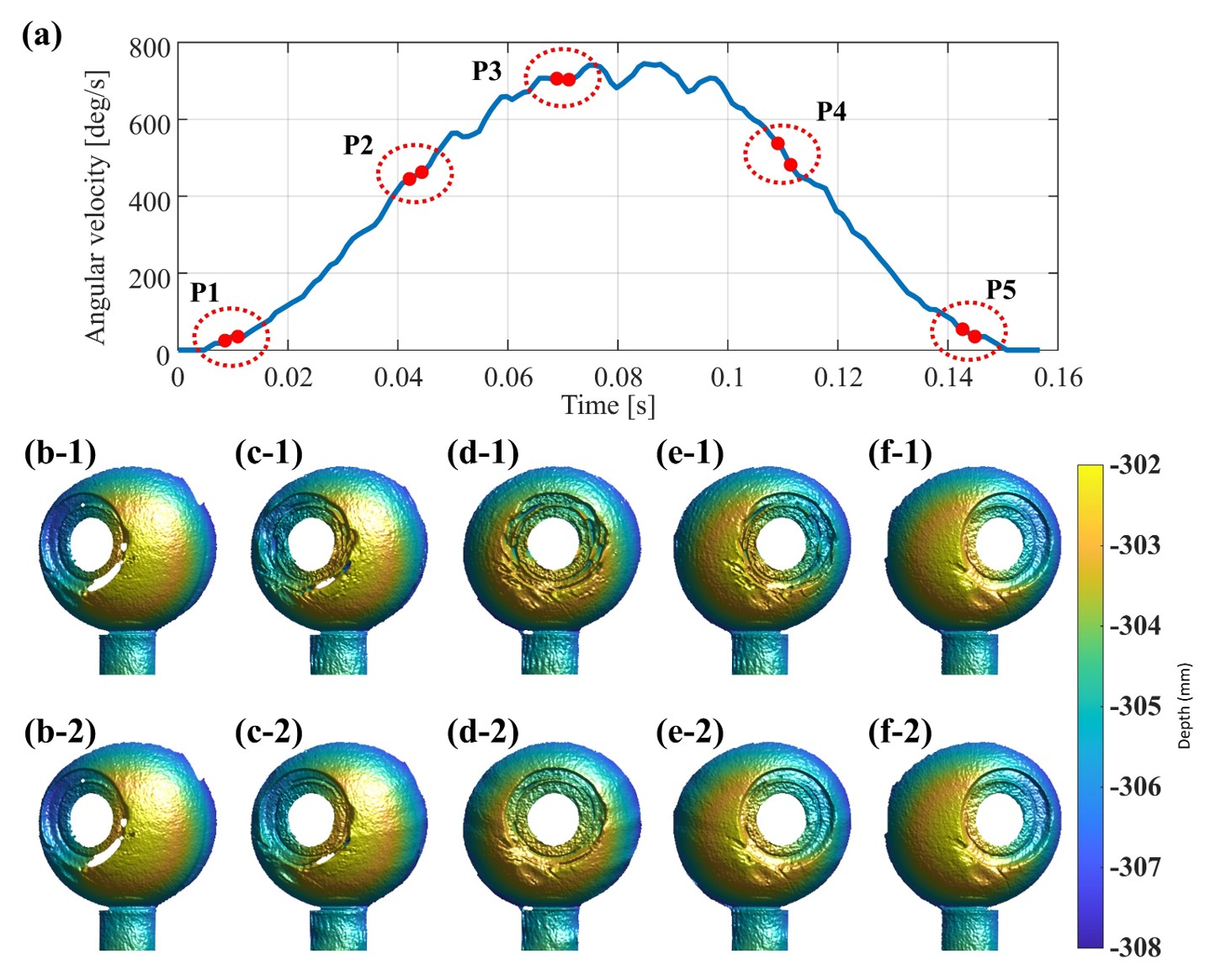}
    \caption{Reconstruction results at selected motion states:(a) angular velocity profile; (b–f) reconstruction results at P1–P5. The first and second rows show the results before and after motion compensation, respectively.}
    \label{fig:pose_results_500fps}
\end{figure}

The compensation performance was further examined at 500 FPS for different eye poses along the rotational trajectory. As shown in Fig.~\ref{fig:pose_results_500fps}, the uncompensated reconstruction results exhibited more noticeable artifacts as the rotational velocity increased.
 In particular, the frontal pose, where the eye motion was relatively severe, showed pronounced deformation in the iris region before compensation. In contrast, after compensation, the iris region was reconstructed more stably for all tested poses, including the side and frontal poses. Based on these observations, the frontal pose with severe motion-induced errors was selected as the main target condition for the subsequent evaluation.

Although the 500 FPS acquisition condition provided reliable reconstruction under fast eye motion, such high-speed imaging hardware may be difficult to implement in practical VR/AR mobile environments due to constraints on system size, power consumption, heat generation, and cost. Therefore, the proposed method was also evaluated under various effective frame-rate conditions. In addition to the actual datasets acquired at 500, 400, 300, and 200 FPS, lower-frame-rate conditions were simulated by temporal downsampling. Specifically, 125 FPS and 100 FPS datasets were generated by sampling every fourth frame from the 500 FPS and 400 FPS datasets, respectively, and a 60 FPS dataset was generated by sampling every seventh frame from the 420 FPS dataset. This downsampling strategy allowed the compensation performance at lower effective frame rates to be examined while excluding additional motion blur effects.

\begin{figure*}[t]
    \centering
    \includegraphics[width=\textwidth]{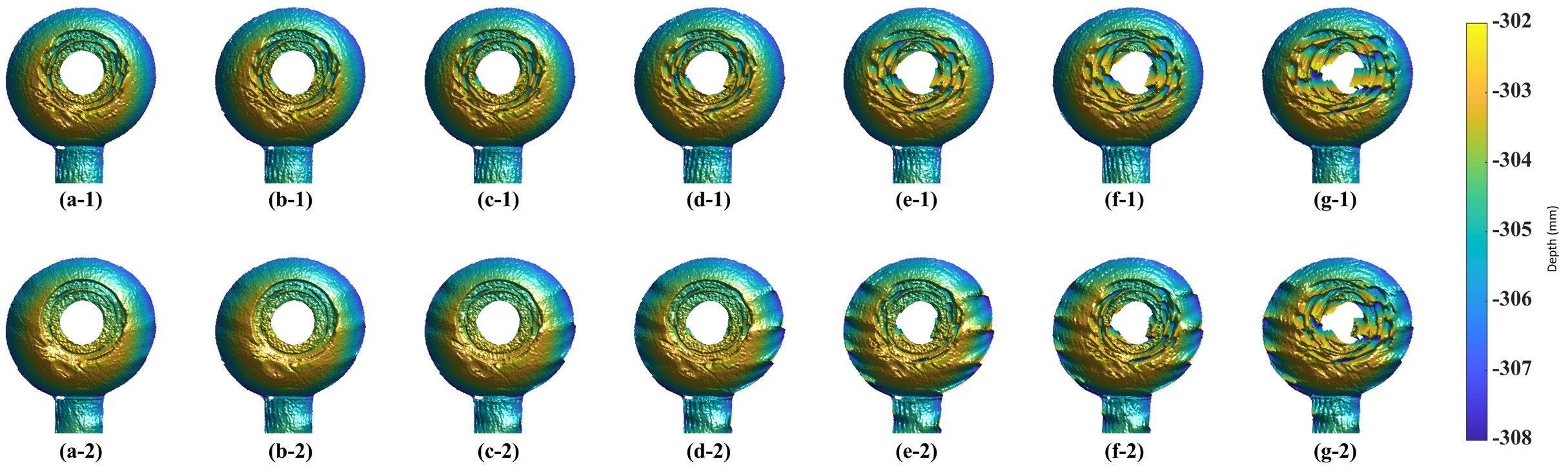}
    \caption{Comparison of 3D reconstruction results at different frame rates: (a) 500 FPS, (b) 400 FPS, (c) 300 FPS, (d) 200 FPS, (e) 125 FPS, (f) 100 FPS, and (g) 60 FPS. The first and second rows show the results before and after motion-induced error compensation, respectively.}
    \label{fig:fps_comparison}
\end{figure*}

Fig.~\ref{fig:fps_comparison} compares the reconstruction results before and after motion compensation under different frame-rate conditions. The proposed motion compensation effectively reduced the dominant motion-induced deformation in the iris region at high and moderate frame rates. However, the compensation effect was not uniform over the entire ocular surface when the effective frame rate became low. Under low-frame-rate conditions, such as 100 FPS and 60 FPS, residual ripple-like artifacts became more pronounced, especially in the sclera region. This indicates that motion compensation alone can reduce the main iris deformation but may also amplify local residual errors when the temporal interval between phase-shifted images becomes large.

This limitation is attributed to the increasing mismatch between the estimated and actual inter-frame motion at lower effective frame rates. Small errors in the estimated eyeball center, pupil position, or rotational motion parameters can produce larger compensation errors as the time interval between phase-shifted images increases. Therefore, although the proposed motion compensation model is effective for suppressing the dominant rotational distortion, additional refinement is required to improve the global reconstruction accuracy under low-frame-rate acquisition conditions.

\subsection{Region-Wise Optimization}
As discussed in the previous section, the proposed motion compensation method reduced the dominant motion-induced deformation in the iris region. However, residual ripple-like artifacts were still observed, especially in the sclera region, as the effective frame rate decreased. This indicates that a single compensation scale applied to the entire ocular surface is not sufficient when the temporal interval between phase-shifted images becomes large. Since the iris and sclera regions have different geometric and optical characteristics, the same motion compensation strength can produce different residual errors in each region.

To address this issue, region-wise optimization was applied by separately refining the compensation strength for the iris and sclera regions. The iris region was segmented by projecting a circular iris boundary, constructed from the estimated eye center, pupil center, surface normal, and predefined iris radius, onto the reference image plane. The remaining valid ocular surface was assigned to the sclera region, while a narrow boundary band was treated separately to avoid abrupt discontinuities between the two regions. Each region was independently compared with a pose-aligned 500 FPS compensated reconstruction, which was used as the high-frame-rate reference. The scale factors applied to the estimated compensation amount were then adjusted to minimize the reconstruction difference from the reference. Finally, the optimized iris and sclera regions were merged to obtain the final region-wise optimized 3D reconstruction.

\begin{figure*}[t]
    \centering
    \includegraphics[width=0.8\textwidth]{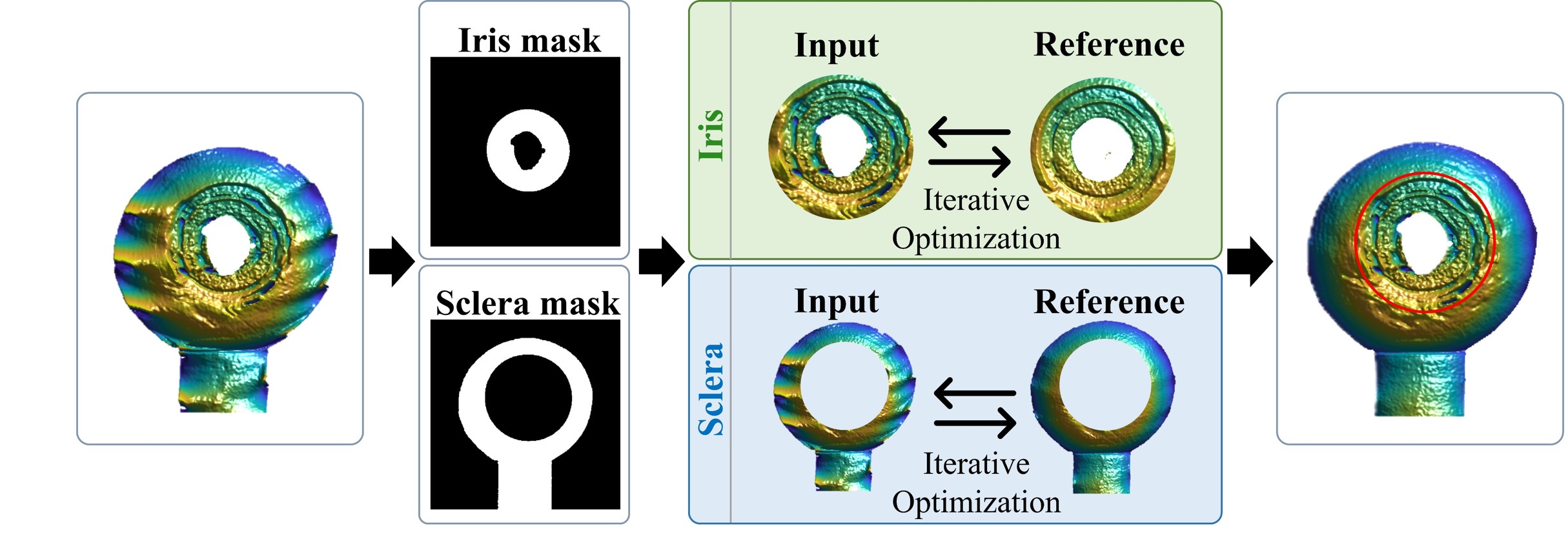}
    \caption{Schematic illustration of the region-wise optimization.}
    \label{fig:region-wise_optimization}
\end{figure*}

\begin{figure*}[t]
    \centering
    \includegraphics[width=\textwidth]{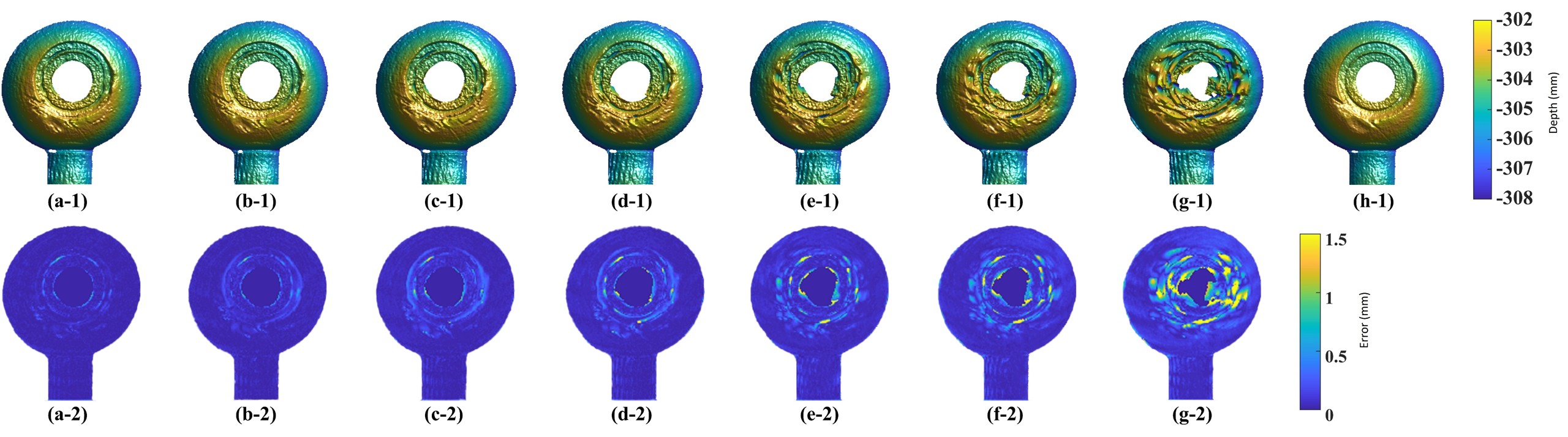}
    \caption{3D reconstruction results after region-wise optimization and comparison with the static GT: (a) 500 FPS, (b) 400 FPS, (c) 300 FPS, (d) 200 FPS, (e) 125 FPS, (f) 100 FPS, and (g) 60 FPS. The first and second rows show the reconstructed results and error maps, respectively. (h) Static GT.}
    \label{fig:optimization_results}
\end{figure*}

Fig.~\ref{fig:optimization_results} shows the reconstruction result after applying region-wise optimization, the static ground-truth 3D reconstruction, and the corresponding error map. Compared with the result obtained using motion compensation alone, the ripple-like artifacts in the sclera region were substantially reduced after region-wise optimization. In addition, the iris region maintained a stable reconstructed shape, indicating that the proposed optimization strategy can improve local reconstruction quality without degrading the region where motion compensation was already effective.

The quantitative evaluation was performed by comparing each reconstructed result with the static ground-truth 3D scan after ICP alignment. The RMSE values for different methods and frame-rate conditions are summarized in Table~\ref{tab:rmse_static_gt}. Compared with the conventional three-step PSP method, the proposed method with motion compensation and region-wise optimization reduced the average RMSE from 0.3660 mm to 0.2136 mm, corresponding to a 41.64\% reduction. This result demonstrates that the proposed framework effectively improves the overall reconstruction accuracy under dynamic eye rotation.

The benefit of region-wise optimization was particularly evident at low frame rates. Although motion compensation reduced the RMSE under most frame-rate conditions, its improvement became less pronounced as the effective frame rate decreased. This is because larger inter-frame eye motion at lower frame rates caused residual ripple-like artifacts, especially in the sclera region. At 60 FPS, motion compensation alone even resulted in a higher RMSE than the conventional three-step PSP method, indicating that motion compensation alone is not sufficient under very low effective frame-rate conditions. After applying region-wise optimization, these residual artifacts were effectively suppressed, and the RMSE decreased across all tested frame-rate conditions compared with the motion-compensation-only results. These results indicate that region-wise optimization compensates for local residual mismatch caused by imperfect motion estimation and improves the robustness of 3D eye reconstruction under low-frame-rate acquisition conditions.

\begin{table*}[t]
\centering
\caption{RMSE(mm) comparison of reconstructed 3D results after ICP alignment with static GT.}
\label{tab:rmse_static_gt}
\begin{tabular}{lccccccc}
\toprule
\multirow{2}{*}{Method}
& \multicolumn{7}{c}{Frame Rates}
\\
\cmidrule(lr){2-8}
& 500 FPS & 400 FPS & 300 FPS & 200 FPS
& 125 FPS & 100 FPS & 60 FPS \\
\midrule
Conventional 3-step PSP
& 0.1763 & 0.1999 & 0.2613 & 0.3338
& 0.4481 & 0.4769 & 0.6654 \\

Motion Compensation
& 0.0760 & 0.0913 & 0.1209 & 0.1982
& 0.3686 & 0.4271 & 0.7214 \\

Motion Compensation + RWO
& \textbf{0.0642} & \textbf{0.0812} & \textbf{0.1191} & \textbf{0.1559}
& \textbf{0.2444} & \textbf{0.2909} & \textbf{0.5392} \\
\bottomrule
\end{tabular}

\vspace{2mm}
\footnotesize{Values are reported in millimeters. RWO denotes the proposed region-wise optimization.}
\end{table*}

\subsection{Additional Validation Using a Non-Spherical Rigid Object}

To further examine whether the proposed compensation strategy is limited to the fake-eye geometry, an additional experiment was performed using a non-spherical rigid object. The object consisted of a planar base with raised square and triangular features, which introduced sharp boundaries, local depth discontinuities, and non-spherical surface geometry. Since the pupil-based motion cue was not available for this object, the motion was estimated from the image-plane displacement of the square marker.

\begin{figure*}[t]
    \centering
    \includegraphics[width=0.9\textwidth]{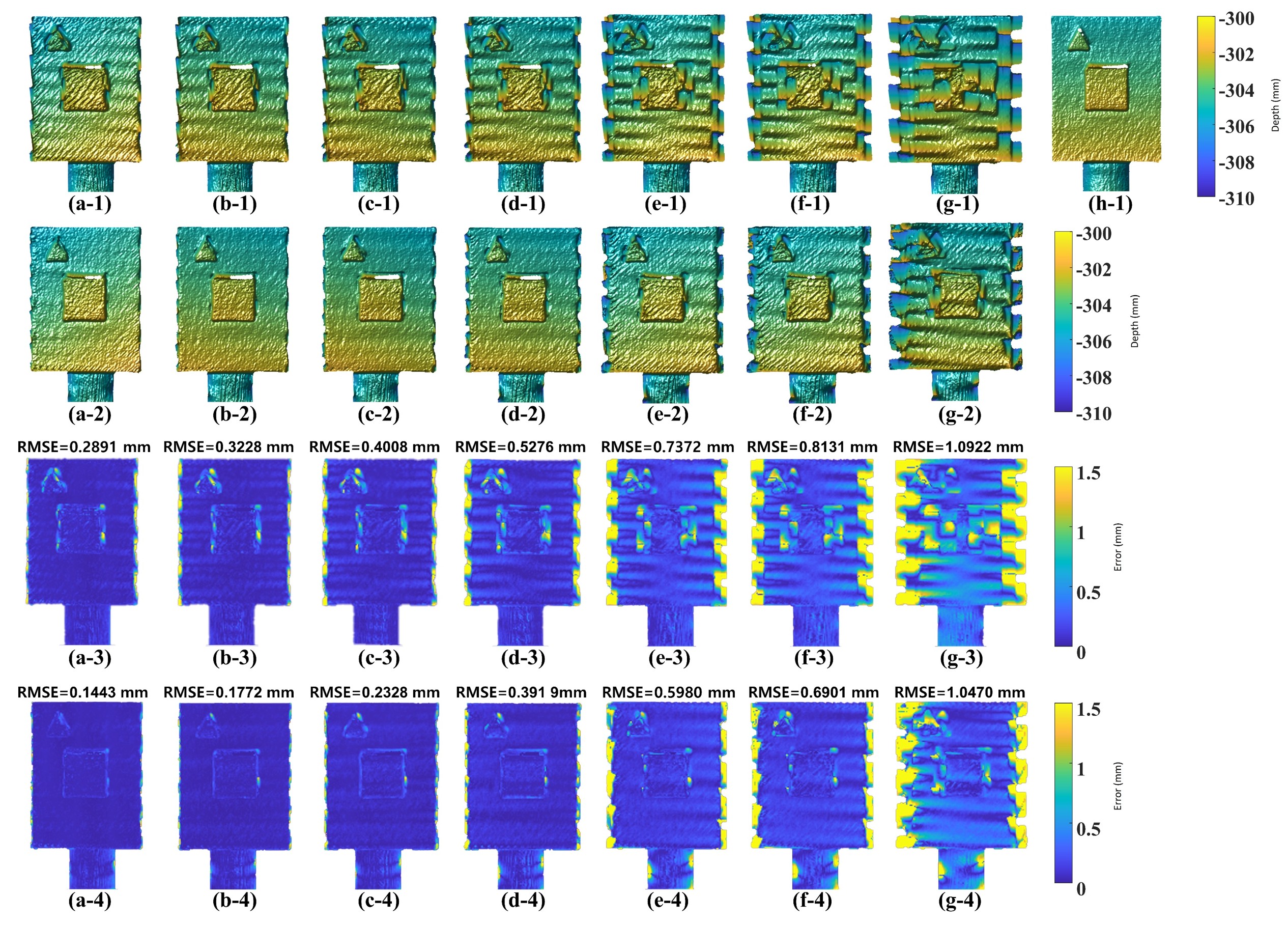}
    \caption{Comparison of 3D reconstruction results for the non-spherical object at different frame rates:
    (a) 500 FPS, (b) 400 FPS, (c) 300 FPS, (d) 200 FPS, (e) 125 FPS, (f) 100 FPS, and (g) 60 FPS.
    The first row shows the results reconstructed using the conventional three-step PSP method, and the second row shows the results reconstructed using the proposed method.
    The third and fourth rows present the corresponding error maps with respect to the static ground-truth 3D shape.
    (h) Static ground-truth 3D shape.}
    \label{fig:optimization_results_square}
\end{figure*}

As shown in Fig.~\ref{fig:optimization_results_square}, the uncompensated reconstruction exhibited noticeable artifacts near the object boundaries and raised features because the phase-shifted images were captured at different rotational states. After applying the proposed motion compensation, the overall surface deformation was reduced, and the reconstructed shape became more stable. The optimized result further suppressed residual errors over most of the object surface, although relatively large errors remained near sharp boundaries due to occlusion, discontinuity, and interpolation effects.

These results suggest that the proposed compensation principle is not strictly limited to the spherical fake-eye geometry. Rather, it can also reduce motion-induced PSP errors for other rigid objects undergoing rotational motion, provided that reliable image features or markers are available for estimating inter-frame motion. However, this experiment should be interpreted as an additional validation of the compensation principle rather than as the primary target application of the proposed framework.

\section{Conclusion}

This study proposed a rotational motion-induced error compensation method for phase-shifting profilometry-based 3D eye reconstruction. Although three-step PSP can provide dense and accurate 3D surface information with a minimal number of projected patterns, sequential image acquisition is vulnerable to eye rotation, which causes inter-frame pixel mismatch and phase-shift errors. To address this issue, the proposed method estimated the rotational motion of the eye in a spherical-coordinate domain and compensated for both camera-pixel and phase-shift errors between phase-shifted fringe images.

The experimental results demonstrated that the proposed compensation method effectively reduced motion-induced deformation in the reconstructed eye surface under high-speed rotational motion. In particular, the iris region, which is critical for reliable 3D gaze-vector estimation, was reconstructed more stably after compensation. The method was further evaluated under various frame-rate conditions, including both actual high-speed acquisition and temporally downsampled datasets. Although residual ripple-like artifacts became more noticeable at lower effective frame rates, especially in the sclera region, the proposed region-wise optimization further improved the reconstruction quality by separately refining the compensation strength for the iris and sclera regions.

Compared with conventional three-step PSP, the proposed framework consistently improved the reconstruction accuracy across the tested frame-rate conditions. These results indicate that rotational motion compensation combined with region-wise optimization can enhance the robustness of PSP-based 3D eye reconstruction in dynamic eye-motion scenarios. Therefore, the proposed method provides a useful technical basis for high-precision 3D eye tracking in immersive VR/AR environments, where stable gaze estimation is required under rapid and non-uniform eye movements.

Despite these promising results, several limitations remain to be addressed. The present experiment was conducted using a fake eye without the corneal part in order to isolate the effect of rotational motion-induced errors. Therefore, additional factors such as corneal refraction, tear-film reflection, eyelid or eyelash occlusion, and individual anatomical variations should be considered in future studies using real human eyes. In addition, the proposed method assumes a fixed eyeball center during reconstruction. In practical HMD usage, however, head turning or device slippage may cause dynamic changes in the relative position between the eye and the sensor, requiring adaptive estimation of the eyeball center. Future work will focus on extending the proposed framework to real human-eye experiments, improving its robustness under lower-frame-rate acquisition, and implementing a compact HMD-compatible structured-light system. Beyond VR/AR applications, the proposed approach may also be extended to other fields that require precise three-dimensional eye tracking, such as ocular proton therapy.

An additional experiment using a non-spherical rigid object further showed that the proposed compensation principle can also reduce motion-induced reconstruction errors for general rigid objects with sharp boundaries and depth discontinuities.

\printcredits

\section*{Declaration of competing interest}
The authors declare that they have no known competing financial interests or personal relationships that could have appeared to influence the work reported in this paper.

\section*{Acknowledgments}
This research was supported by the Technology Innovation Program(Project Name: Development of AI autonomous continuous production system technology for gas turbine blade maintenance and regeneration for power generation, Project Number: RS-2025-25447257, Contribution Rate: 30\%) funded By the Ministry of Trade, Industry and Resources(MOTIR, Korea), the Culture, Sports and Tourism R\&D Program through the Korea Creative Content Agency grant funded by the Ministry of Culture, Sports and Tourism in 2024 (Project Name: Global Talent for Generative AI Copyright Infringement and Copyright Theft, Project Number: RS-2024-00398413, Contribution Rate: 30\%), and the National Research Foundation of Korea(NRF) grant funded by the Korea government(MSIT) (Project Number: RS-2026-25505230, Contribution Rate: 30\% / Project Number: RS-2026-25498577, Contribution Rate: 10\%).

\section*{Data Availability}
Data will be made available on request.

\bibliographystyle{cas-model2-names}
\bibliography{cas-refs}

\end{document}